\documentclass[conference]{IEEEtran}
\IEEEoverridecommandlockouts

\usepackage{cite}
\usepackage{amsmath,amssymb,amsfonts}
\usepackage{algorithmic}
\usepackage{graphicx}
\usepackage{textcomp}
\usepackage{xcolor}
\usepackage{booktabs}
\usepackage{url}

\def\BibTeX{{\rm B\kern-.05em{\sc i\kern-.025em b}\kern-.08em
    T\kern-.1667em\lower.7ex\hbox{E}\kern-.125emX}}
\begin{document}

\title{ShipwreckFinder: A QGIS Tool for Shipwreck Detection in Multibeam Sonar Data
\thanks{$^*$Denotes equal contribution.}
\thanks{Corresponding Author Email: \texttt{anjashep@umich.edu}}
\thanks{This work was supported by the NOAA Ocean Exploration program under Award \#NA23OAR0110315 and by the National Science Foundation under Award \#DGE2241144.}
}

\author{
\IEEEauthorblockN{Anja Sheppard$^{1^*}$,
Tyler Smithline$^{1^*}$,
Andrew Scheffer$^{2}$,
David Smith$^{2}$,\\
Advaith V. Sethuraman$^{1}$,
Ryan Bird$^{1}$,
Sabrina Lin$^{3}$,
Katherine A. Skinner$^{1}$}
\IEEEauthorblockA{$^{1}$Department of Robotics\\
$^{2}$Department of Electrical Engineering and Computer Science\\
$^{3}$Computer Science in the College of Literature, Science, and the Arts\\
University of Michigan, Ann Arbor, MI, USA}}

\maketitle

\begin{abstract}
In this paper, we introduce ShipwreckFinder, an open-source QGIS plugin that detects shipwrecks from multibeam sonar data. Shipwrecks are an important historical marker of maritime history, and can be discovered through manual inspection of bathymetric data. However, this is a time-consuming process and often requires expert analysis. Our proposed tool allows users to automatically preprocess bathymetry data, perform deep learning inference, threshold model outputs, and produce either pixel-wise segmentation masks or bounding boxes of predicted shipwrecks. The backbone of this open-source tool is a deep learning model, which is trained on a variety of shipwreck data from the Great Lakes and the coasts of Ireland. Additionally, we employ synthetic data generation in order to increase the size and diversity of our dataset. We demonstrate superior segmentation performance with our open-source tool and training pipeline as compared to a deep learning-based ArcGIS toolkit and a more classical inverse sinkhole detection method. The open-source tool can be found at \url{https://github.com/umfieldrobotics/ShipwreckFinderQGISPlugin}.
\end{abstract}

\begin{IEEEkeywords}
marine perception, shipwreck segmentation, deep learning, QGIS
\end{IEEEkeywords}

\section{Introduction}
Recent advances in acoustic sensor technology and marine survey platforms have enabled efficient large area data collection to deliver massive amounts of data to marine scientists. For example, high-resolution mapping projects such as Lakebed 2030 \cite{glos} aim to fully map the seafloor of the Great Lakes in the next half decade, greatly increasing the amount of publicly available data. However, this data has yet to be fully leveraged for training machine learning models.

Sunken objects such as shipwrecks and airplanes hold important archaeological, historical, and environmental data. Finding shipwrecks in large-area seafloor surveys is a time-consuming task. Typically, this is done by hand with experts who manually inspect statistical anomalies in the data and cross-reference historical shipwreck logs \cite{plets2011}. In the past decade, interest in algorithmic and deep learning approaches to shipwreck detection has increased. A variety of sensors have been used for this problem, ranging from Multibeam Echosounder (MBES) \cite{davis2020, pols2025, arcgis} and backscatter data \cite{masetti2012}, to orbital ocean imagery taken from satellites \cite{baeye2016}, to sidescan sonar \cite{sethuraman2023}, to bathymetry from LiDaR \cite{character2021}. However, these existing approaches often perform poorly on out-of-distribution data, still require expert oversight, and are rarely open-source. 

\begin{figure}
    \centering
    \includegraphics[width=\linewidth]{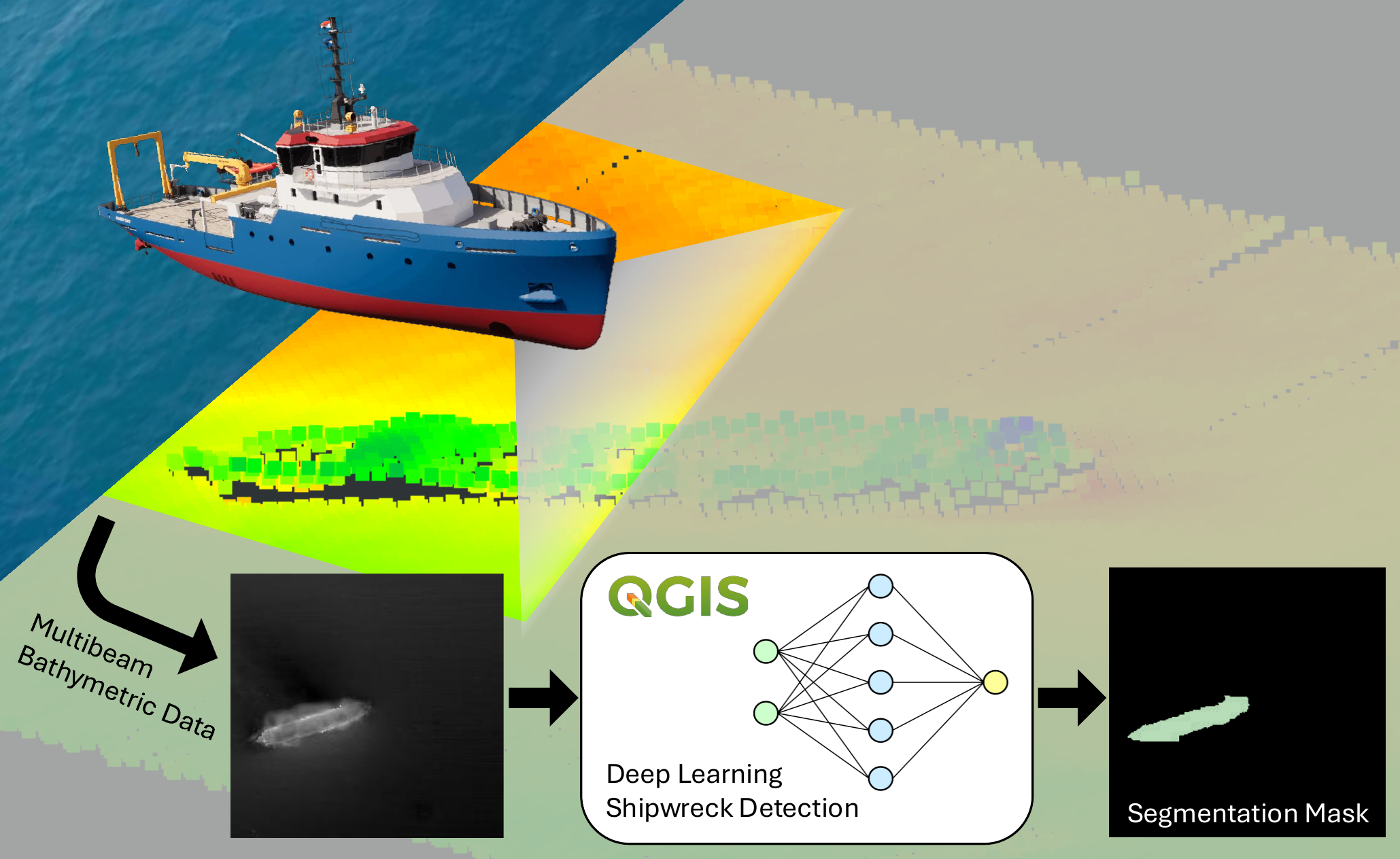}
    \caption{Our proposed QGIS plugin, ShipwreckFinder, takes a multibeam sonar scan as input and returns a segmentation mask of predicted shipwrecks.}
    \label{fig:overview}\vspace{-1em}
\end{figure}

In this work, we develop a machine learning-based tool to automate the detection of shipwreck sites from multibeam sonar data (see Fig. \ref{fig:overview}). Our tool, ShipwreckFinder, is designed to have seamless integration into QGIS \cite{QGIS_software}, a freely available Geographic Information System (GIS) platform, to enable visualization and geo-referencing for detected shipwrecks. We train and validate a shipwreck segmentation model using existing data from Thunder Bay National Marine Sanctuary (TBNMS) \cite{sethuraman2024} and deeper water data collected by the National Oceanic and Atmospheric Administration (NOAA) \cite{ncei} and Integrated Mapping for the Sustainable Development of Ireland's Marine Resource (INFOMAR) \cite{infomar_shipwrecks}. ShipwreckFinder aims to improve access to state-of-the-art machine learning methods within the marine archaeology community through an open-source toolset for automatic shipwreck detection. 

Ultimately, the ShipwreckFinder tool has great potential to reduce the time and cost required to detect archaeological sites from multibeam sonar data collected across our lakes and oceans, accelerating the timeline for discoveries to be made by the scientific community and shared with the public.

\renewcommand{\arraystretch}{1.2}
\begin{table*}[ht]
    \centering
    \caption{Comparison of existing shipwreck detection methods}
    \label{tab:methods}
    \begin{tabular}{c|ccccc}
        \toprule
        Method && Sensor Modality & Deep Learning & Predicts Segmentation Masks & Open-Source Tool \\
        \midrule
        \midrule
        Plets et al. \cite{plets2011} && MBES & & & \\
        Masetti et al. \cite{masetti2012} && MBES Backscatter & & \checkmark & \\
        Davis et al. \cite{davis2020} && MBES  & & \checkmark & \checkmark \\
        Singh et al. \cite{arcgis} && MBES & \checkmark & \checkmark & \checkmark \\
        Character et al. \cite{character2021} && MBES, LiDAR & \checkmark & \\
        Sethuraman et al. \cite{sethuraman2023} && SSS & \checkmark & \checkmark &  \\
        Ard et al. \cite{ard2023} && MBES + SSS & \checkmark & \checkmark & \\
        Pols et al. \cite{pols2025} && MBES & \checkmark & & \\
        Ours && MBES & \checkmark & \checkmark & \checkmark \\
        \bottomrule
    \end{tabular}
\end{table*}
\renewcommand{\arraystretch}{1.0}

\section{Related Works}

\subsection{Deep Learning Approaches to Segmentation}

Deep learning approaches have become popular and commonplace for tasks such as object detection and segmentation. 
Convolutional Neural Networks (CNNs) such as U-Net \cite{ronneberger2015} use a symmetric encoder-decoder structure with skip connections to learn segmentation prediction masks from a labeled training dataset. HRNet \cite{wang2020} does not use a traditional encoder-decoder structure, but performs segmentation while maintaining a high-resolution representation of the data. Salient Object Detection (SOD) approaches such as BASNet \cite{qin2019} aim to learn the boundaries of the ``salient" object that stands out from its surroundings, such as a shipwreck on terrain.

Although there is an abundance of negative (non-shipwreck) data publicly available, there remain few examples of shipwrecks for the models to learn from. This makes the use of data-hungry models such as transformers more challenging to train. This is often the case in field robotics applications where data collection is time-consuming and expensive. One potential solution is to supplement the training data with simulated images. STARS leveraged simulated data for shipwreck detection on sidescan sonar data \cite{sethuraman2023}. To train ShipwreckFinder, we construct an augmented training set with real shipwrecks randomly superimposed on a variety of additional terrain to increase the amount of training data available for shipwreck detection from multibeam sonar data.

\subsection{Bathymetric Shipwreck Detection}

Several approaches to shipwreck detection from sonar data exist in the literature \cite{masetti2012, baeye2016, davis2020, arcgis, character2021, ard2023, pols2025}. Manual inspection is a common approach to finding shipwrecks, although time-consuming \cite{plets2011}. In an effort to use automated approaches to shipwreck detection, both hand-crafted \cite{masetti2012, baeye2016, davis2020} and learning-based approaches have been proposed \cite{character2021, arcgis, ard2023, pols2025}.  An early method \cite{masetti2012} uses a classical approach by analyzing the multibeam sonar backscatter returns for shipwreck-identifying features, which are differentiated from non-shipwreck terrain using clustering. The use of satellite imagery rather than sonar data has also been proposed \cite{baeye2016}, but it only works in shallow turbid conditions. Another approach \cite{davis2020} uses the built-in ArcGIS depression analysis tool on inverted bathymetry scans. In \cite{character2021}, a YOLOv3 model is trained on labeled bathymetric data to detect shipwreck bounding boxes. Another proposed approach involves compositing side-scan sonar (SSS) and multibeam sonar data into an input image for a segmentation model \cite{ard2023}. A very recent work presents a mix of classical and learning-based approaches by analyzing topographic signatures \cite{pols2025}. Most relevant to our work, the ArcGIS Deep Learning module has also been used for shipwreck detection. However, the ArcGIS model requires a software license to use and requires the user to label existing data in ArcGIS to train the model \cite{arcgis}.

In this work, we propose a shipwreck detection toolset that includes a pre-trained backbone for shipwreck segmentation. Our tool is available as a plugin in QGIS, making it broadly available and open-source. We compare our method against the two other existing methods that use MBES data, predict segmentation masks, and have an open-source tool \cite{davis2020, arcgis}.

\section{Technical Approach}
\begin{figure*}
    \centering
    \includegraphics[width=\linewidth]{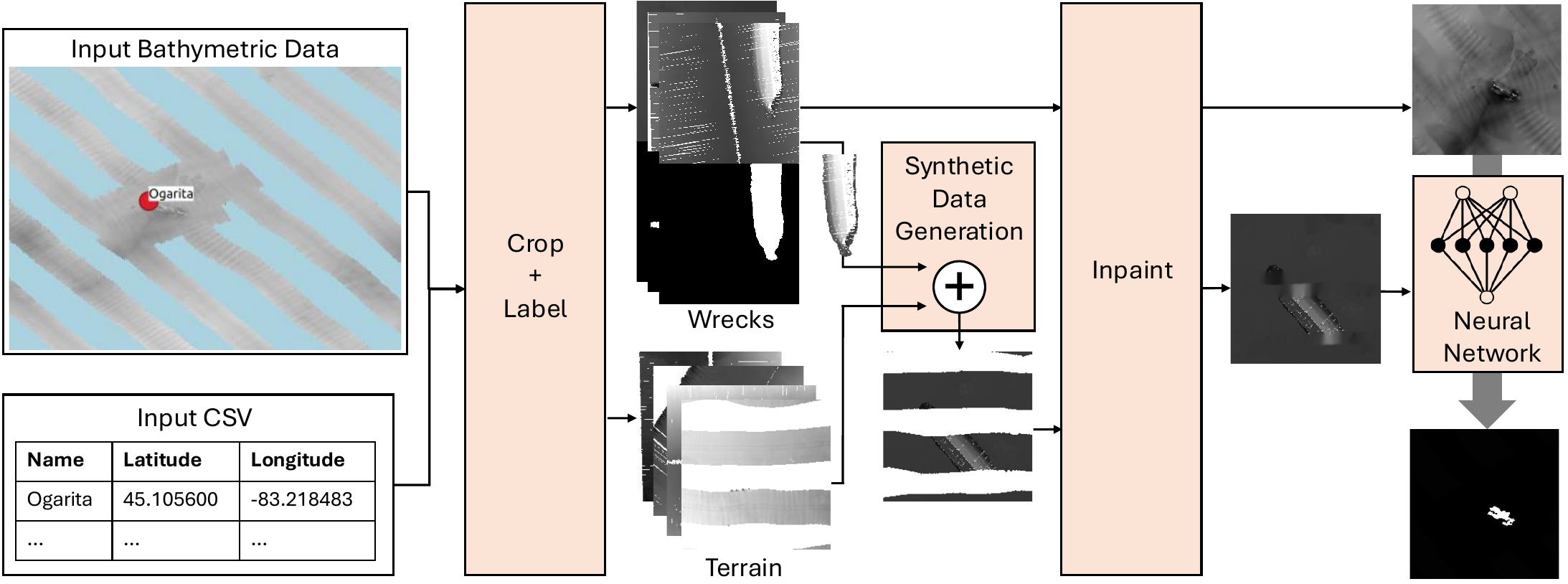}
    \caption{An overview of our data processing pipeline for training the shipwreck detection model. We cross-reference known ship locations with raw BAG data and crop out 200 x 200 meter sections. Then we manually label the wrecks, and pass these wreck images and labels in addition to blank terrain into a randomized synthetic data generation module, which produces additional training data. Finally, we use a mix of real and synthetic data to train the model.}
    \label{fig:data_process}
\end{figure*}

In this section, we provide background for the multibeam sonar sensor model and discuss the technical details of our shipwreck segmentation model and the development of the QGIS plugin. An overview of our approach is illustrated in Fig. \ref{fig:data_process}.

\subsection{Multibeam Sonar Sensor Model}

The multibeam sonar sensor emits a fan-shaped array of acoustic beams that span a defined swath perpendicular to the vehicle’s trajectory. The acoustic beams propagate through the water medium, and a portion of the beam is backscattered upon encountering a surface, such as the seafloor or a shipwreck. The time-of-flight and intensity information received by the sensor is used to construct a 3D point cloud \cite{thurman2009}. Data is collected as a vehicle is moving, taking motion compensation into account, in order to construct the scene.

This differs from SSS data, which constructs a waterfall image based off of returns from dual fan-shaped acoustic beams, and from forward-looking sonar, which produces a full image of a shorter-range scene at a high frequency.

Bathymetric data is typically collected with a sensor mounted to the hull of a large vessel, but can also be collected with an Autonomous Underwater Vehicle (AUV).

\subsection{Dataset Preparation: Crop and Label}
To train our model, we use data from NOAA Ocean Exploration's publicly available survey data \cite{noaa_ncei_bathy}, the INFOMAR database \cite{infomar_shipwrecks}, and field data collected from our prior field expeditions in TBNMS \cite{sethuraman2024}. We compile a total of 60 data files from these three sources. We then manually label each shipwreck image using existing SSS labels as a reference \cite{sethuraman2024}, which is fairly trivial given the small dataset size. The surveys used to compose our dataset are detailed on the project webpage.

We work with postprocessed data products in Bathymetric Attributed Grid (BAG), XYZ, and TIFF formats. Using the metadata, we check known shipwreck coordinates against the boundaries of each survey \cite{infomar_shipwrecks}. If the coordinate is within the boundaries, a cropped grid is extracted. If any data points within the extracted grid are nonzero, and therefore contain data, then the grid is saved for use in the dataset.

As each data product has varying resolution, spanning from 0.5 m/pixel to 10 m/pixel, we crop each extracted grid to the same spatial resolution of $200 \times 200$ meters. This results in a dataset with images of varying sizes in pixel space. Additionally, all training data is stored as binary \texttt{numpy} files rather than PNGs in order to retain depth information.

This process yields 162 grids containing 58 unique shipwrecks.

\begin{figure*}[t]
    \centering
    \includegraphics[width=\linewidth]{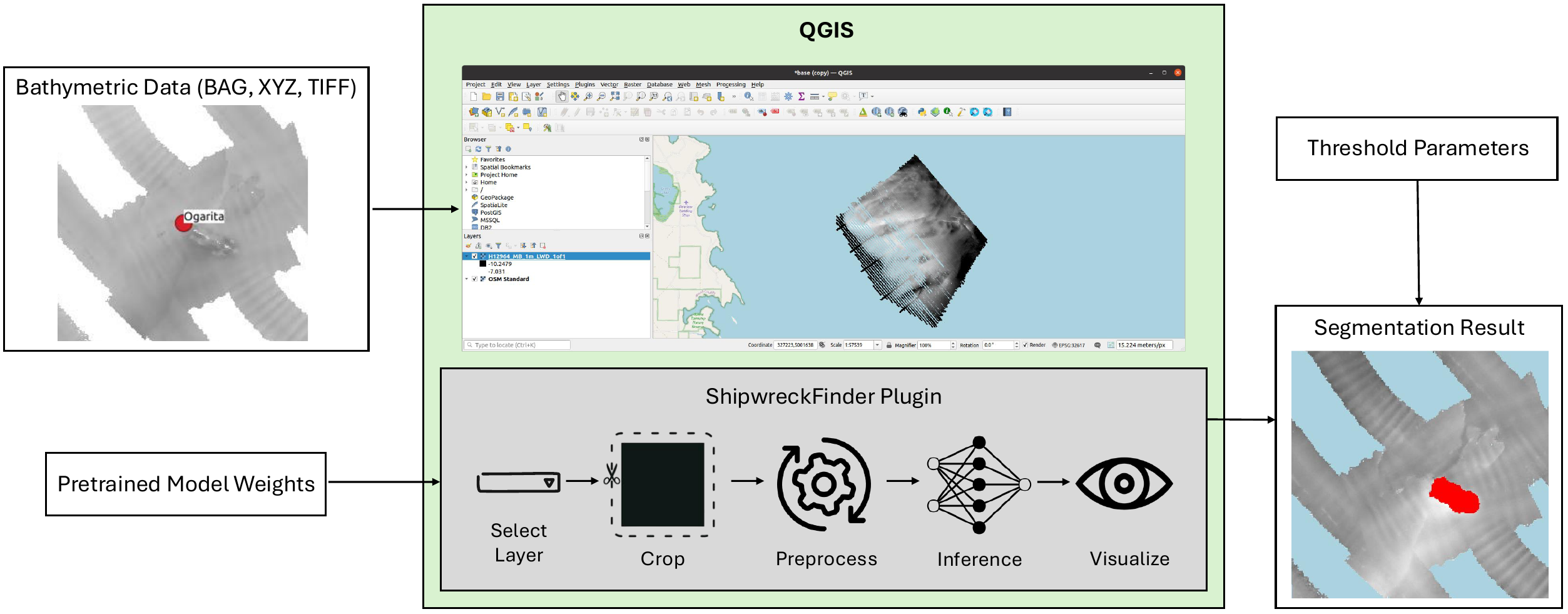}
    \caption{An overview of the workflow for our proposed ShipwreckFinder QGIS plugin. The user's bathymetric data is passed through a preprocessing step and then our trained model in order to output either pixel-wise segmentation or bounding boxes, which are derived from the segmentation output.}
    \label{fig:workflow}
\end{figure*}
 
\subsection{Dataset Preparation: Synthetic Data Generation}
In order to increase the diversity and size of our training set, we employ a synthetic data generation pipeline. We accomplish this by extracting ships from the cropped grids based off of their label, and then compositing them at a random location and orientation onto a terrain grid.

The composited ships and terrain are not guaranteed to lie at the same depths. To account for this, we first calculate the mean depth of ships and terrain from their source images. For ships, the mean depth is averaged over all pixels contained within the ``ship" label. For terrain, the mean depth is averaged over all valid pixels in the cropped grid. The ship is then composited onto the terrain data such that the mean ship depth is normally distributed around 91\% of the mean terrain depth. This produces a dataset that closely reflects the composition of the original dataset. An accompanying pixel-wise segmentation label is also automatically computed based off the composited ship location and rotation.

After synthetic data generation, the full resulting dataset has 1784 images, with a total of 162 real shipwreck images, 455 synthetic composited shipwreck images, and 1167 terrain images. The dataset is split into 65\% train images, 5\% validation images, and 30\% test images.

\subsection{Dataset Preparation: Inpaint}
Bathymetric data is often spatially incomplete, with data gaps, holes, or streaks. These streaks are not uniform, such as with the nadir in SSS, making them difficult for the network to adapt to. In order to address this, we use OpenCV Navier-Stokes inpainting with a radius of 8 pixels in order to avoid confusing the network near survey edges \cite{bertalmio2001}.

\subsection{Neural Network Model Architecture}
We explore several model architectures for shipwreck segmentation, including U-Net \cite{ronneberger2015}, U-Net-Hillshade, HRNet \cite{wang2020}, and Salient Object Detection (SOD) \cite{qin2019}.

\subsubsection{U-Net} 

The U-Net architecture \cite{ronneberger2015} is a well-established and highly effective model for supervised image segmentation. In our implementation, we use a ResNet-34 network as the encoder to extract key hierarchical features from the input image. The encoder begins with a $7\times7$ convolution with 64 filters, followed by four downsampling stages composed of residual blocks with increasing depth. Intermediate feature maps from each stage are stored and used as skip connections to retain spatial detail. The decoder consists of a series of five upsampling blocks that iteratively increase the resolution, employing skip connections and upconvolutions to produce a segmentation map that matches the input resolution.

\subsubsection{U-Net-Hillshade}

In our modified U-Net-Hillshade model, we augment the U-Net model input by incorporating a hillshade representation of the image patch as an additional channel. Hillshade imagery enhances the visibility of shipwreck boundaries, potentially facilitating improved feature extraction and enabling the model to more effectively learn spatial patterns associated with shipwrecks.

\subsubsection{HRNet}

HRNet \cite{wang2020} was originally designed to learn from high-resolution image representations. Although our dataset contains a range of resolutions, including low-resolution images, the parallelized multi-resolution approach proposed by HRNet serves as an alternative to the traditional encoder-decoder framework present in U-Net. We specifically employ the HRNet-Object Contextual Representations (OCR) implementation, which has parallel branches at four resolutions: 1/4, 1/8, 1/16, and 1/32. Additionally, the OCR module uses object-level context to refine the segmentation mask via an attention mechanism.

\subsubsection{Salient Object Detection}

The relative infrequency of shipwrecks present in large-area bathymetric surveys makes the segmentation problem extremely unbalanced. One potential alternative approach is treating shipwrecks as the ``salient object" against a terrain background. We train an SOD network, BASNet \cite{qin2019}, which utilizes a predict-refine architecture to produce highly accurate boundary predictions. Additionally, this model is designed for mobile and web use, making it highly efficient to run during inference time.

\begin{table*}[!ht]
    \centering
    \caption{Model Training Parameters}
    \label{table:model-training}
    \begin{tabular}{l|cccccc}
        \toprule
        Model & \# Epochs & Starting Learning Rate & Learning Rate Scheduler & Optimizer & Loss Function & Batch Size \\
        \midrule
        \midrule
        U-Net & 12,000 & 5e-4 & OneCycle & Adam & Cross Entropy & 64 \\
        U-Net-Hillshade & 12,000 & 5e-4 & OneCycle & Adam & Cross Entropy & 64 \\
        HRNet & 12,000 & 7e-4 & Plateau & Adam & Cross Entropy & 64 \\
        BASNet & 4,300 & 3e-4 & Plateau & Adam & Cross Entropy + & 16\\
        & & & & & Structural Similarity + IoU &  \\
        \bottomrule
    \end{tabular}
\end{table*}

\begin{table*}[!ht]
    \centering
    \caption{Segmentation Accuracy Comparison}
    \label{table:iou}
    \begin{tabular}{l|ccccccc}
        \toprule
        && $\text{IoU}_\text{ship}$ ($\uparrow$) & $\text{IoU}_\text{terrain}$ ($\uparrow$)& F1 Score ($\uparrow$) & Precision ($\uparrow$)& Recall ($\uparrow$) & Wreck Count Percentage ($\uparrow$) \\
        \midrule
        \midrule
        Davis et al. \cite{davis2020} && 0.139 &	0.941 & 0.243 & 0.321 & 0.196 & 0.447 \\
        Singh et al. \cite{arcgis} && 0.005 & 0.953 & 0.001 & \textbf{0.881} & 0.005 & 0.000\\
        \midrule
        Ours: U-Net && 0.176 & 0.945 & 0.299 & 0.481 & 0.218 & 0.426 \\
        Ours: U-Net-Hillshade && \textbf{0.494} & \textbf{0.962} & \textbf{0.661} & 0.649 & \textbf{0.674} & 0.681 \\
        Ours: HRNet && 0.375 & 0.945 & 0.546 & 0.502 & 0.597 & \textbf{0.787} \\
        Ours: BASNet && 0.331 & 0.961 & 0.497 & 0.825 &  0.356 & 0.723 \\
        \bottomrule
    \end{tabular}
\end{table*}

\subsection{Plugin Development}
We implemented ShipwreckFinder as an open-source plugin for QGIS to enable shipwreck detection directly within a familiar and accessible GIS environment. The plugin is designed to be modular, user-friendly, and compatible with a variety of bathymetric data formats, including BAG, XYZ, and TIFF files. All core components are written in Python, leveraging the QGIS Python API \cite{QGIS_software}.

The ShipwreckFinder workflow is shown in Fig. \ref{fig:workflow}, and is detailed below. The plugin allows the user to select an entire bathymetric layer, or draw an extent over a layer for more targeted predictions. All preprocessing, inference, and post-processing occurs in one window, making the tool easy to use for QGIS users. In order to support a variety of machines, from desktops with GPUs to laptops in the field, the plugin also intelligently batches and re-merges inference results in order to avoid overloading computer memory. 
The output is a segmentation layer, which overlays the original extent and identifies the predicted shipwreck pixels. Additionally, there is an option to output bounding boxes, derived from the segmentation predictions, depending on the user's downstream task needs. After the user selects the layer, the plugin goes through two main processing steps, with an optional third step to produce bounding boxes:

\subsubsection{Preprocessing}
The preprocessing stage converts input bathymetric rasters into several $200 \times 200$ meter cropped ``chunks" for input to the network. Each chunk is then normalized and infilled with the OpenCV Navier-Stokes infilling function.

\subsubsection{Inference}
For model inference, each processed chunk is passed through the selected model weights. Segmentation outputs are reconstructed back into a continuous layer to display the output, and the threshold parameter specified by the user is applied to remove contours smaller than the specified size (in order to remove small, noisy predictions). To reduce memory usage, if there are more than 500 chunks being merged, they are recursively stitched together in batches until only one continuous layer remains.

\subsubsection{Bounding Box Extraction}
The bounding box extraction step transforms thresholded segmentation masks into polygonal vector layers, producing bounding boxes around detected structures. These vectorized outputs can be exported, queried, or used in downstream spatial workflows within QGIS.

\section{Experiments and Results}
We present experiments highlighting the strengths and weaknesses of our ShipwreckFinder plugin, including a comparison against two baseline methods \cite{davis2020, arcgis}, an analysis of the four model architectures, a demonstration of the plugin's computation time, and an investigation of the shipwreck parameters impacting the model performance. Additionally, we provide details about our model training parameters.

\subsection{Field Dataset Collection}

The additional field data used for model training was collected using the Michigan Technological University (MTU) Iver3 AUV equipped with an EdgeTech 2205 dual frequency 540/1610 kHz SSS and 3D bathymetric system \cite{sethuraman2024}. The surveys were pre-programmed either in a lawnmower or an object identification pattern. As the primary goal of these surveys was to collect SSS data, the resulting bathymetric data is non-standard for typical processed surveys. Instead of a large mosaic of MBES data collected over a wide area, each ship is closely inspected and the data is not combined within a site. This results in several wreck sites having multiple surveys and imaging angles.

\subsection{Model Training}
Each of the trained models is detailed in Table \ref{table:model-training}. Training was performed on a system with an AMD EPYC 7742 64-core CPU, running Ubuntu 22.04, equipped with an NVIDIA A100-SXM4 GPU (80GB VRAM) and CUDA 12.8. Identical training, test, and validation sets are used to train and evaluate each of our models, with the exception of U-Net-Hillshade which also requires hillshades of the training set as an input.

\subsection{Baseline Methods}

We compare our approach against two baseline methods: an inverse sinkhole detection method \cite{davis2020} and a pretrained shipwreck detection model \cite{arcgis}. Both methods are run in ArcGIS, in order to most closely match the original papers.

\subsubsection{Inverse Sinkhole Detection}

For this approach, we run three steps on each raster: First, we invert the raster. Then, we use a custom tool \cite{davis2020} to detect sinkholes, where we set the \texttt{MinDepress} parameter to 100 and the \texttt{Buffer} parameter to 1. The third step is a final depression analysis filtering tool \cite{davis2020}, which isolates potential wrecks from the identified sinkholes. We set \texttt{Interval} = 0.2, \texttt{MinDepth} = 0.2, and the \texttt{Base} to the raster minimum depth rounded to one decimal point. For the timing experiment in Section \ref{sec:runtime}, we sum the processing time for each of these steps for each layer. As this is an analytical method, there was no training process. However, we did use wrecks and their corresponding labels from the train set to assist in manually tuning the parameter values.

\subsubsection{ArcGIS Pretrained Shipwreck Detection}

The pretrained shipwreck detection model \cite{arcgis} is trained on a small subsection of Jamaica Bay in New York. We did not re-train the model on our data, as we wanted to evaluate the utility of the tool as released to the community. 
For parameters, we kept the suggested \texttt{threshold} to 0.4 after manually evaluating several threshold values on examples from the train set.

\begin{figure*}
    \centering
    \includegraphics[width=0.97\linewidth]{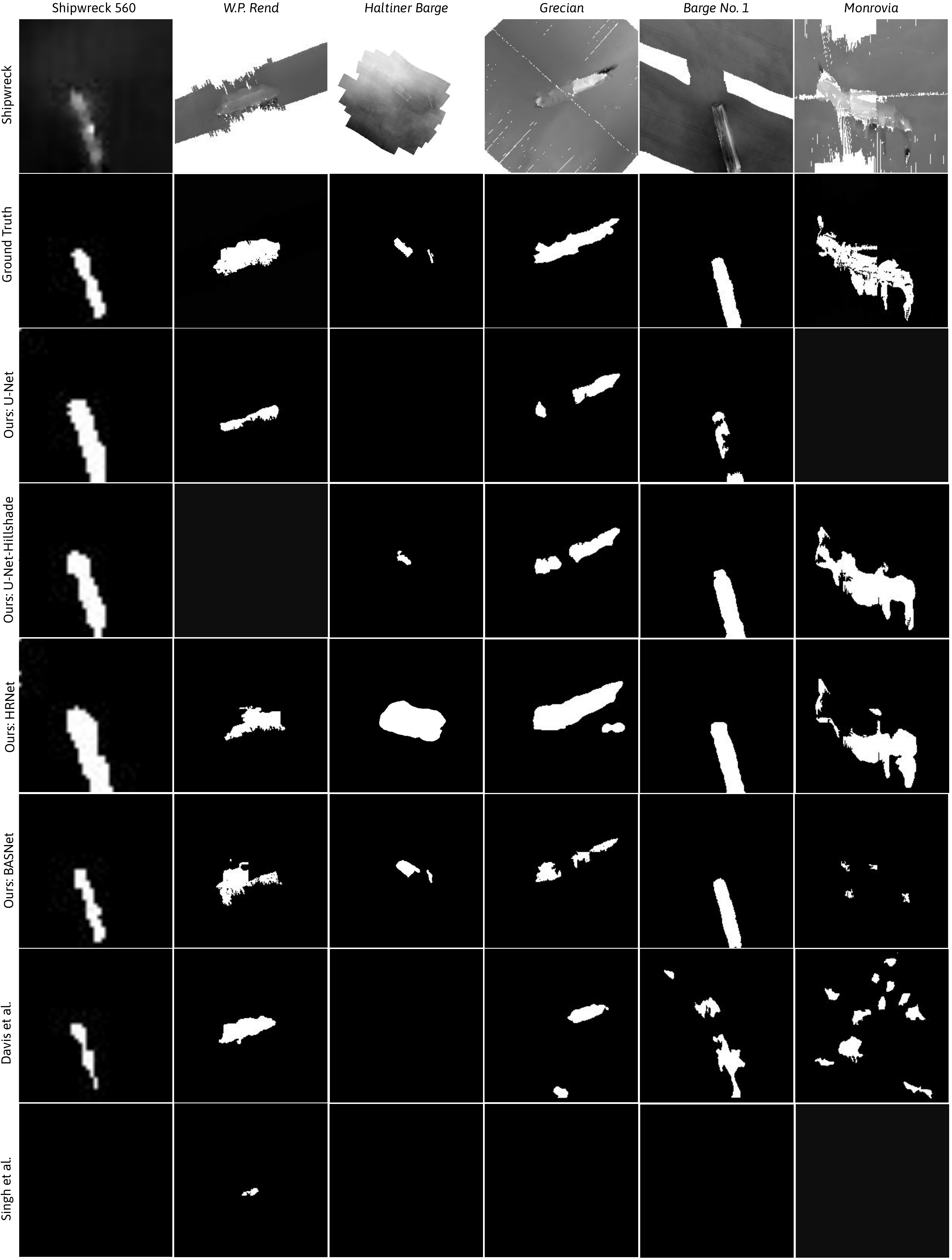}
    \caption{Qualitative comparison of the two baseline methods \cite{davis2020, arcgis} and our methods on several real wrecks. On most of the test sites, the ArcGIS method \cite{arcgis} does not produce any predictions.}
    \label{fig:qual}
\end{figure*}

\begin{figure*}[!ht]
    \centering
    \includegraphics[width=\textwidth]{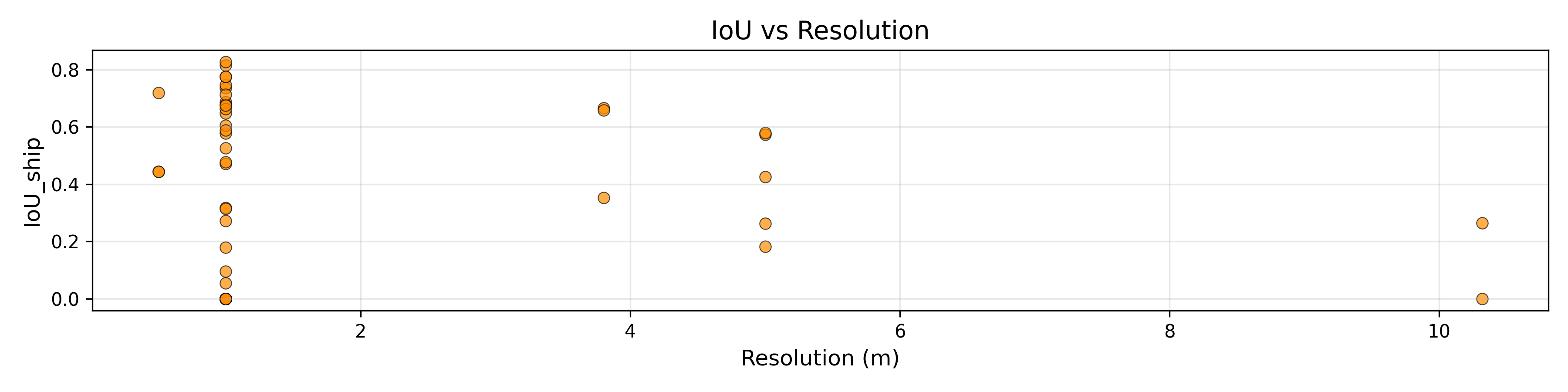}
    \vspace{1em} 
    \includegraphics[width=\textwidth]{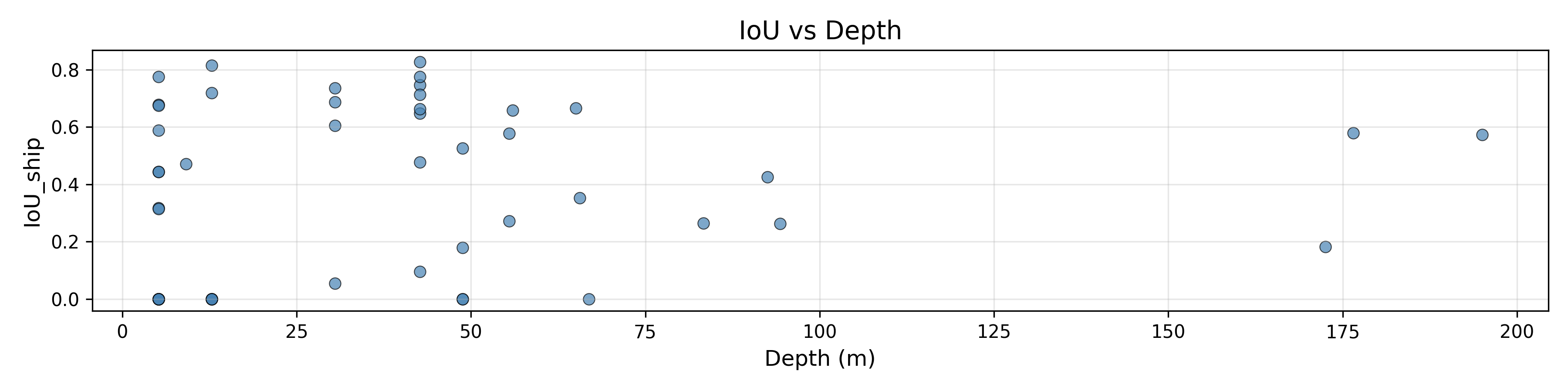}
    \caption{Top: U-Net-Hillshade IoU$_\text{ship}$ vs Resolution. Bottom: U-Net-Hillshade IoU$_\text{ship}$ vs Depth.}
    \label{fig:stacked_images}
    \vspace{-1em}
\end{figure*}

\subsection{Segmentation Accuracy}

We compare the segmentation accuracy of our model against baseline approaches \cite{davis2020, arcgis} in Table~\ref{table:iou}. We use per-class Intersection over Union (IoU), F1 score, precision, and recall as metrics, as defined by the PASCAL VOC challenge \cite{everingham2010}. All metrics are calculated directly from the total true positive, false positive, true negative, and false negative pixel counts across the entire test set, making higher resolution BAG files have a higher weight in the final IoU. We additionally show the wreck count percentage, which displays the percentage of wrecks that have an IoU$_\text{ship}$ of at least 0.2. This aims to demonstrate which method is best at identifying the existence of the most wrecks correctly, even if the segmentation mask may not be the most accurate. We see HRNet excels and has the highest rate of shipwreck identification, while U-Net-Hillshade has the best overall segmentation accuracy with the highest IoU.

We show qualitative results with all four of our deep learning architectures as compared to the two baselines on six wrecks in Fig. \ref{fig:qual}.

\subsection{Architecture Comparison}

We also evaluate several different segmentation architectures in order to explore which model type is able to detect shipwrecks with the highest accuracy. The results are displayed in Table \ref{table:iou}. The majority of the metrics calculated from our test set indicate that our proposed modified architecture U-Net-Hillshade is the best performing model, although all four tested models outperform both baselines.

In Fig. \ref{fig:qual}, we display qualitative results from the four different architectures. Note that BASNet seems to have the most accurate fine-grained segmentation, but at the cost of many more false negatives (\textit{Monrovia} is almost completely missed). Due to the superior IoU demonstrated by U-Net-Hillshade, we use this model for the next two experiments on runtime and depth. However, HRNet showed the best performance for the wreck count percentage metric. Since each model architecture has different strengths, they will all be available as potential options in the plugin for the user to choose from.

\subsection{Plugin Runtime}\label{sec:runtime}

In this experiment, we compare the runtimes of our approach and the baseline approaches. As each layer is a different size, we report the runtime per MB by computing the ratio of the runtime $r$ (in seconds) with the size of the layer $s$ (in MB) and then averaging across all $n$ layers:
\begin{equation}
    R = \frac{1}{n} \sum_{i=1}^{n} \frac{r_i}{s_i}
\end{equation}
This results in the runtime results as displayed below in Table \ref{table:run}. We show runtime results for the best-performing method, U-Net-Hillshade, on both GPU and CPU. Even if the user does not have a GPU for running inference, the speed at which our plugin runs vastly outperforms the two baselines. 

\begin{table}[!ht]
    \centering
    \caption{Runtime Comparison}
    \label{table:run}
    \begin{tabular}{lc}
        \toprule
        & Runtime per MB (s/MB) \\
        \midrule
        \midrule
        Davis et al. \cite{davis2020} (CPU) & 251.22 \\
        Singh et al. \cite{arcgis} (GPU) & 172.61 \\
        Ours: U-Net-Hillshade (GPU) & \textbf{2.95} \\
        Ours: U-Net-Hillshade (CPU) & 3.62 \\
        \bottomrule
    \end{tabular}
\end{table}

\subsection{Accuracy vs Depth}

In this experiment, we explore the impact of the data resolution and shipwreck depth on the model segmentation performance. See Fig. \ref{fig:stacked_images} for plots showing IoU$_\text{ship}$ versus depth and resolution. It seems that wreck depth does not have a major impact on model performance, but data resolution may have a slight negative relationship to IoU. This is important to note as most of the MBES survey coverage that is open-source is relatively low resolution for the task of shipwreck detection, so acquiring more high-resolution survey data will be important to identifying more wrecks.

\section{Conclusion}

In this work, we present ShipwreckFinder, an open-source deep learning-based plugin for segmenting shipwrecks from bathymetric data. We detail the training process, which involved compiling a dataset from available data in the Great Lakes and the Irish coast and generating synthetic data to augment the real dataset. Our model outperforms two baseline methods that use both hand-crafted and learning-based approaches for this task. Additionally, we conduct several experiments to better understand the strengths and limitations of ShipwreckFinder. The plugin will be open-sourced in order to allow for broad use by the marine archaeology and underwater robotics communities.

\section*{Acknowledgments}


We acknowledge the lives that were lost in shipwrecks throughout TBNMS and the coasts of Ireland. This work was supported by the NOAA Ocean Exploration program under Award \#NA23OAR0110315 and by the National Science Foundation under Award \#DGE2241144.

\bibliographystyle{ieeetr}
\bibliography{references}

\begin{thebibliography}{10}

\bibitem{glos}
{Great Lakes Observing System}, ``Lakebed 2030,'' 2025.

\bibitem{plets2011}
R.~Plets, R.~Quinn, W.~Forsythe, K.~Westley, T.~Bell, S.~Benetti, F.~McGrath, and R.~Robinson, ``Using multibeam echo-sounder data to identify shipwreck sites: Archaeological assessment of the joint irish bathymetric survey data,'' {\em International Journal of Nautical Archaeology}, vol.~40, no.~1, pp.~87--98, 2011.

\bibitem{davis2020}
D.~S. Davis, D.~C. Buffa, and A.~C. Wrobleski, ``Assessing the utility of open-access bathymetric data for shipwreck detection in the united states,'' {\em Heritage}, vol.~3, no.~2, pp.~364--383, 2020.

\bibitem{pols2025}
C.~Pols, F.~Sturt, C.~El~Safadi, and A.~Marcu, ``Shipwreck detection using semi-automated methods: Combining machine learning and topographic inference approaches,'' {\em Journal of Archaeological Science}, 2025.

\bibitem{arcgis}
R.~Singh and V.~Viswambharam, ``How we did it: Detecting shipwrecks using deep learning at {UC} 2020.'' ArcGIS Blog, 2020.

\bibitem{masetti2012}
G.~Masetti and B.~Calder, ``Remote identification of a shipwreck site from {MBES} backscatter,'' {\em Journal of Environmental Management}, vol.~111, pp.~44--52, 2012.

\bibitem{baeye2016}
M.~Baeye, R.~Quinn, S.~Deleu, and M.~Fettweis, ``Detection of shipwrecks in ocean colour satellite imagery,'' {\em Journal of Archaeological Science}, vol.~66, pp.~1--6, 2016.

\bibitem{sethuraman2023}
A.~V. Sethuraman and K.~A. Skinner, ``{STARS}: Zero-shot sim-to-real transfer for segmentation of shipwrecks in sonar imagery,'' in {\em Proceedings of the British Machine Vision Conference 2023}, 2023.

\bibitem{character2021}
L.~Character, A.~Ortiz~Jr, T.~Beach, and S.~Luzzadder-Beach, ``Archaeologic machine learning for shipwreck detection using lidar and sonar,'' {\em Remote Sensing}, vol.~13, no.~9, p.~1759, 2021.

\bibitem{QGIS_software}
{QGIS Development Team}, {\em QGIS Geographic Information System}.
\newblock QGIS Association, 2025.

\bibitem{sethuraman2024}
A.~V. Sethuraman, A.~Sheppard, O.~Bagoren, C.~Pinnow, J.~Anderson, T.~C. Havens, and K.~A. Skinner, ``Machine learning for shipwreck segmentation from side scan sonar imagery: Dataset and benchmark,'' {\em The International Journal of Robotics Research}, vol.~44, no.~3, pp.~341--354, 2024.

\bibitem{ncei}
``{NOAA National Centers for Environmental Information}.'' \url{https://www.ngdc.noaa.gov/nos}.
\newblock Accessed January 2025.

\bibitem{infomar_shipwrecks}
{Geological Survey Ireland}, ``{INFOMAR Shipwrecks Dataset},'' 2023.

\bibitem{ard2023}
W.~Ard and C.~Barbalata, ``Sonar image composition for semantic segmentation using machine learning,'' in {\em Proceedings of the IEEE/CVF Winter Conference on Applications of Computer Vision}, pp.~248--254, 2023.

\bibitem{ronneberger2015}
O.~Ronneberger, P.~Fischer, and T.~Brox, ``{U-Net}: Convolutional networks for biomedical image segmentation,'' in {\em International Conference on Medical image computing and computer-assisted intervention}, pp.~234--241, Springer, 2015.

\bibitem{wang2020}
J.~Wang, K.~Sun, T.~Cheng, B.~Jiang, C.~Deng, Y.~Zhao, D.~Liu, Y.~Mu, M.~Tan, X.~Wang, W.~Liu, and B.~Xiao, ``Deep high-resolution representation learning for visual recognition,'' {\em IEEE Transactions on Pattern Analysis and Machine Intelligence}, vol.~43, no.~10, pp.~3349--3364, 2020.

\bibitem{qin2019}
X.~Qin, Z.~Zhang, C.~Huang, C.~Gao, M.~Dehghan, and M.~Jagersand, ``{BASNet}: Boundary-aware salient object detection,'' in {\em The IEEE Conference on Computer Vision and Pattern Recognition (CVPR)}, June 2019.

\bibitem{thurman2009}
E.~Thurman, J.~Riordan, and D.~Toal, ``Multi-sonar integration and the advent of sensor intelligence,'' in {\em Advances in Sonar Technology} (S.~R. Silva, ed.), ch.~7, Rijeka: IntechOpen, 2009.

\bibitem{noaa_ncei_bathy}
{NOAA National Centers for Environmental Information}, ``Bathymetric data viewer.'' \url{https://www.ncei.noaa.gov/maps/bathymetry/?layers=multibeam}, 2025.
\newblock Accessed 2025.

\bibitem{bertalmio2001}
M.~Bertalmio, A.~Bertozzi, and G.~Sapiro, ``Navier-stokes, fluid dynamics, and image and video inpainting,'' in {\em Proceedings of the 2001 IEEE Computer Society Conference on Computer Vision and Pattern Recognition. CVPR 2001}, vol.~1, pp.~I--I, 2001.

\bibitem{everingham2010}
M.~Everingham, L.~Van~Gool, C.~K. Williams, J.~Winn, and A.~Zisserman, ``The {PASCAL} visual object classes ({VOC}) challenge,'' {\em International Journal of Computer Vision}, vol.~88, no.~2, pp.~303--338, 2010.

\end{thebibliography}



\end{document}